\documentclass{article}

\usepackage{arxiv}

\usepackage[utf8]{inputenc} 
\usepackage[T1]{fontenc}    
\usepackage[citecolor=blue]{hyperref}       
\usepackage{url}            
\usepackage{booktabs}       
\usepackage{amsfonts}       
\usepackage{nicefrac}       
\usepackage{microtype}      
\usepackage{lipsum}		
\usepackage{graphicx}
\usepackage{natbib}
\usepackage{doi}

\usepackage{cite}
\usepackage{amsmath,amssymb,amsfonts}
\usepackage{algorithmic}
\usepackage{graphicx}
\usepackage{algorithm,algorithmic}
\hypersetup{hidelinks=true}
\usepackage{textcomp}

\usepackage{graphics} 
\usepackage{epsfig} 
\usepackage{mathptmx} 
\usepackage{times} 
\usepackage{balance}

\usepackage{subcaption}
\usepackage[T1]{fontenc}
\usepackage{booktabs}
\usepackage{makecell}

\usepackage{float}
\usepackage{longtable}
\usepackage{rotating}
\usepackage{multicol}
\usepackage{bm}
\usepackage{multirow}
\usepackage{colortbl}
\usepackage{xcolor}
\usepackage{graphicx}
\usepackage{array}
\usepackage{makecell}
\usepackage{tabularx}

\newcolumntype{L}[1]{>{\raggedright\arraybackslash}m{#1}}
\newcolumntype{C}[1]{>{\centering\arraybackslash}m{#1}}
\newcolumntype{Y}{>{\centering\arraybackslash}X}

\def\BibTeX{{\rm B\kern-.05em{\sc i\kern-.025em b}\kern-.08em
    T\kern-.1667em\lower.7ex\hbox{E}\kern-.125emX}}
\begin{document}
\title{FOUND-AF: Benchmarking ECG Foundation Models for Atrial Fibrillation Detection}

\author{ \href{https://orcid.org/0009-0009-8349-0239}{\includegraphics[scale=0.06]{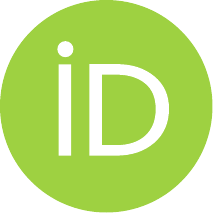}\hspace{1mm}Amirhossein~Taleshinosrati}\\
	SDU Health Informatics and Technology\\
	University of Southern Denmark\\
	Odense, 5230, Denmark \\
	\texttt{amtal25@student.sdu.dk} \\
	\And
	\href{https://orcid.org/0009-0003-6142-2442}{\includegraphics[scale=0.06]{orcid.pdf}\hspace{1mm}Yangyang~Wang} \\
	Department of Computer Science\\
	University of Helsinki\\
	Helsinki, 00100, Finland \\
	\texttt{wang.yangyang@helsinki.fi} \\
	\And
	\href{https://orcid.org/0009-0000-8284-110X}{\includegraphics[scale=0.06]{orcid.pdf}\hspace{1mm}Atitaya~Phoemsuk} \\
	School of Computer Science and Electronic Engineering\\
	University of Essex\\
	Colchester, CO4 3SQ, UK\\
	\texttt{ap19698@essex.ac.uk} \\
	\And
	\href{https://orcid.org/0000-0002-2151-5180}{\includegraphics[scale=0.06]{orcid.pdf}\hspace{1mm}Vahid~Abolghasemi} \\
	School of Computer Science and Electronic Engineering\\
	University of Essex\\
	Colchester, CO4 3SQ, UK\\
	\texttt{v.abolghasemi@essex.ac.uk} \\
	\And
	\href{https://orcid.org/0000-0001-9923-9879}{\includegraphics[scale=0.06]{orcid.pdf}\hspace{1mm}Naser~Hossein~Motlagh} \\
	Department of Computer Science\\
	University of Helsinki\\
	Helsinki, 00100, Finland \\
	\texttt{naser.motlagh@helsinki.fi} \\
    \And
	\href{https://orcid.org/0000-0001-7564-2612}{\includegraphics[scale=0.06]{orcid.pdf}\hspace{1mm}Sadasivan~Puthusserypady} \\
	Department of Health Technology\\
	Technical University of Denmark\\
	Kgs. Lyngby, 2800, Denmark \\
	\texttt{sapu@dtu.dk} \\
    \And
	\href{https://orcid.org/0000-0003-3716-3201}{\includegraphics[scale=0.06]{orcid.pdf}\hspace{1mm}Daniel~Teichmann} \\
	SDU Health Informatics and Technology\\
	University of Southern Denmark\\
	Odense, 5230, Denmark \\
	\texttt{date@mmmi.sdu.dk} \\
    \And
	\href{https://orcid.org/0000-0002-3180-4365}{\includegraphics[scale=0.06]{orcid.pdf}\hspace{1mm}Abdolrahman~Peimankar} \thanks{Corresponding author} \\
	SDU Health Informatics and Technology\\
	University of Southern Denmark\\
	Odense, 5230, Denmark \\
	\texttt{abpe@mmmi.sdu.dk} \\
}


\maketitle

\begin{abstract}
Atrial fibrillation (AF) is the most common sustained cardiac arrhythmia and is associated with increased risks of stroke, heart failure, and mortality. Recent ECG foundation models offer transferable representations for automated AF detection. However, their relative effectiveness remains unclear because existing studies use different datasets, preprocessing procedures, classifiers, and validation protocols. This study presents FOUND-AF, a unified, leakage-controlled, and deployment-oriented benchmarking framework that evaluates the quality of pretrained ECG representations under identical experimental conditions. Nine publicly available foundation models from five families, including HuBERT-ECG, CLEF, ST-MEM, ECG-JEPA, and ECGFounder, were evaluated across four heterogeneous ECG datasets, namely AFDB, CinC2017, CPSC2021, and LTAFDB. All models were used as frozen feature extractors with standardized preprocessing, model-native resampling, a fixed XGBoost classifier, and recording-level grouped cross-validation. The evaluation included classification metrics, receiver operating characteristic analysis, paired recording-level bootstrap comparisons with Holm correction, embedding-space visualization, and computational efficiency profiling. The ECGFounder model consistently achieved the strongest overall performance across datasets while offering a favorable trade-off between accuracy, model size, inference time, and memory usage. FOUND-AF therefore provides a reproducible framework for selecting ECG foundation models and demonstrates that compact, clinically pretrained encoders can support robust and computationally efficient AF detection across heterogeneous acquisition settings.
\end{abstract}

\keywords{Atrial fibrillation \and edge deployment \and electrocardiogram \and foundation models \and machine learning}

\section{Introduction}
\label{sec:introduction}
Atrial fibrillation (AF) is the most common sustained cardiac arrhythmia in adults and represents a major and growing public-health burden worldwide. Its global prevalence has been estimated at more than 33 million cases and is expected to increase further with population ageing and the rising prevalence of cardiovascular risk factors \citep{van2024esc,chugh2014worldwide}. AF is associated with a substantially increased risk of ischemic stroke, heart failure, cognitive decline, and all-cause mortality \citep{wolf1991atrial,schnabel201550}. Since AF can be paroxysmal and asymptomatic, many episodes remain undetected until severe clinical complications occur. Early and reliable AF detection is therefore essential for timely clinical intervention, risk stratification, and initiation of appropriate treatment.

The electrocardiogram (ECG) remains the primary modality for AF diagnosis in both clinical and ambulatory settings. However, the increasing use of Holter monitors, adhesive ECG patches, bedside telemetry, handheld devices, and smartwatches has resulted in large volumes of ECG recordings that cannot be efficiently reviewed by clinicians alone \citep{perez2019large,lubitz2022detection,tison2018passive}. Automated ECG analysis is therefore needed to support scalable AF screening and continuous rhythm monitoring. In particular, robust AF detection from short ECG segments is important for real-time and edge-based monitoring systems, where decisions must be made under limited computational resources and with minimal dependence on cloud connectivity.

Traditional automated AF detection methods have mainly relied on hand-crafted features extracted from RR-interval irregularity, heart rate variability, P-wave morphology, or frequency-domain characteristics, followed by conventional machine learning classifiers such as support vector machines, decision trees, and ensemble methods \citep{tateno2001automatic,dash2009automatic}. Although these approaches are often computationally efficient and clinically interpretable, their performance depends strongly on accurate R-peak detection, expert-driven feature design, and dataset-specific parameter tuning. Consequently, these models may suffer from limited generalizability when applied to ECG recordings acquired using different devices, sampling frequencies, lead configurations, noise conditions, or patient populations.

Deep learning has reduced the need for manual feature engineering by learning discriminative representations directly from ECG signals. Convolutional neural networks, recurrent neural networks, hybrid architectures, and transformer-based models have achieved strong performance in AF and arrhythmia detection tasks \citep{hannun2019cardiologist,ribeiro2020automatic,attia2019artificial,acharya2017deep,yildirim2018novel,lecun2015deep}. However, these models are typically trained in a fully supervised manner and require large amounts of accurately labelled ECG data. Such annotations are costly, time-consuming, and often unavailable in real-world clinical and wearable monitoring scenarios. Moreover, supervised models trained on one dataset may show degraded performance when transferred to another dataset because of differences in acquisition protocols, signal quality, rhythm prevalence, and population characteristics.

Recently, foundation models have emerged as a promising paradigm for biomedical signal analysis. Inspired by the broader foundation-model paradigm in vision and language processing \citep{bommasani2021opportunities}, ECG foundation models are pretrained on large-scale ECG corpora and can subsequently be reused as general-purpose representation learners for downstream cardiac tasks. Several publicly available ECG foundation models have recently been introduced, including HuBERT-ECG \citep{coppola2024hubert}, CLEF \citep{shu2025clef}, ST-MEM \citep{na2024guiding}, ECG-JEPA \citep{kim2024learning}, and ECGFounder \citep{li2025electrocardiogram}. These models employ different architectures and pretraining strategies and their ability to generate transferable ECG embeddings makes them attractive for AF detection, particularly when only lightweight downstream classifiers are desired.

Despite these advancements, the comparative performance and downstream efficacy of ECG foundation models for AF detection remain poorly understood.
Existing studies typically evaluate individual models using different datasets, preprocessing strategies, segmentation schemes, downstream classifiers, and validation protocols. These differences make it difficult to determine whether observed performance gains are due to the pretrained representations themselves or to inconsistencies in the experimental pipeline. Moreover, many evaluations are limited to a single dataset, which provides limited evidence regarding robustness across heterogeneous ECG acquisition settings. Another important but underexplored aspect is the computational efficiency. For practical AF screening in edge and bedside monitoring systems, model accuracy alone is not sufficient; inference time, memory usage, and model size must also be considered.

To address these limitations, this study presents \textit{FOUND-AF}, a unified and deployment-oriented benchmark of ECG foundation models for AF detection. We evaluate nine publicly available ECG foundation models from five model families, namely HuBERT-ECG, CLEF, ST-MEM, ECG-JEPA, and ECGFounder, across four heterogeneous ECG datasets: AFDB, CinC2017, CPSC2021, and LTAFDB. All models are used as frozen feature extractors under an identical preprocessing and evaluation pipeline. The extracted embeddings are classified using a fixed Extreme Gradient Boosting (XG-
Boost) classifier, and performance is evaluated using recording-level grouped cross-validation to reduce the risk of data leakage. In addition to standard classification metrics, we perform receiver-operating-characteristic (ROC) analysis, paired recording-level bootstrap comparisons with Holm correction, and computational efficiency profiling to assess predictive performance and deployment feasibility.

The main contributions of this work are summarized as follows:

\begin{itemize}
    \item We establish a rigorous benchmark evaluating nine ECG foundation models across four heterogeneous datasets, enabling a controlled, systematic comparison of distinct pretrained representations under an identical experimental protocol.
    
    \item We develop a reproducible, frozen-embedding evaluation pipeline that integrates standardized preprocessing, 10-second segmentation, model-native resampling, patient- or recording-level grouped cross-validation, and a standardized downstream XGBoost classifier.
    
    \item We systematically profile the computational efficiency of these architectures regarding parameter size, inference latency, and memory footprint, providing practical insights into their viability for resource-constrained, edge-based AF screening.

\end{itemize}

A preliminary version of this work, which was limited to three ECG foundation-model families evaluated only on one dataset (AFDB), has been accepted for publication at IEEE EMBC 2026 \citep{peimankar2026comparative}. The present study substantially extends that work through a nine-model, four-dataset benchmark with leakage-controlled validation, statistical comparisons, and computational-efficiency analysis.

The remainder of this paper is organized as follows. Section~\ref{sec:relatedworks} reviews relevant literature on ECG-based AF detection and contemporary ECG foundation models. Section~\ref{sec:materialsandmethods} details the datasets, preprocessing pipeline, and evaluated architectures, while Section~\ref{sec:pipeline} outlines the proposed benchmarking framework. The experimental results, including classification performance, statistical validations, embedding spaces, and computational profiles, are presented in Section~\ref{sec:results}. Section~\ref{sec:discussion} contextualizes the core findings, exploring their clinical implications, deployment feasibility, and limitations. Finally, Section~\ref{sec:conclusion} concludes the paper. 

\section{Related Work}
\label{sec:relatedworks}

\subsection{Traditional ECG-Based AF Detection}

Early automated AF detection methods mainly relied on rhythm irregularity, which is one of the main characteristics of AF. These approaches typically extracted hand-crafted features from RR intervals, heart rate variability, P-wave activity, or frequency-domain representations, followed by conventional machine learning or rule-based classifiers \citep{dash2009automatic,petrenas2015low,peimankar2018ensemble,jahan2022short}. Common descriptors include the coefficient of variation of RR intervals, root-mean-square successive differences, entropy-based measures, and geometric representations such as Lorenz or Poincaré plots \citep{wesselius2021digital}. Low-complexity detectors based on similar rhythm descriptors have also been proposed for long-term ECG monitoring, where computational efficiency and robustness are essential \citep{petrenas2015low}. More recent studies have also investigated short-term ECG representations and Poincaré-based analysis for AF detection, further confirming the diagnostic importance of rhythm irregularity in short ECG segments \citep{islam2025poincare}.

Despite these advantages, traditional AF detectors have several limitations. Their performance often depends on accurate upstream R-peak detection, which can be unreliable in noisy ambulatory or wearable ECG recordings. In addition, thresholds and feature sets are frequently optimized for specific datasets, acquisition devices, or patient cohorts, limiting generalizability across heterogeneous recording conditions. Morphology-based features, such as P-wave absence or fibrillatory-wave characteristics, can improve precision but are sensitive to signal quality, lead configuration, and short recording duration. These limitations are particularly important for single-lead and non-standardized ECG recordings, where noise, motion artifacts, and short observation windows can reduce the reliability of hand-crafted features \citep{fan2018multiscaled,lai2020non,zhang2023variable}.

\subsection{Deep Learning Approaches for AF Detection}

Deep learning has substantially changed automated ECG analysis by enabling models to learn discriminative representations directly from raw or minimally processed ECG signals. One-dimensional convolutional neural networks (1-D CNNs) have demonstrated strong performance in arrhythmia classification and AF screening from short single-lead ECG recordings. For example, Fan et al. proposed a multiscale CNN fusion framework for AF screening from short single-lead ECG recordings \citep{fan2018multiscaled}. Lai et al. further investigated AF screening using a non-standardized patch-based ECG lead combined with a deep learning algorithm, which is particularly relevant for wearable and non-clinical acquisition scenarios \citep{lai2020non}. Dual-channel and densely connected neural networks have also been proposed to improve AF detection from single-lead or variable-duration ECG signals \citep{fang2021dual,zhang2023variable}.

Beyond AF-specific methods, several studies have contributed to the broader development of deep learning for ECG analysis. Xu et al. investigated end-to-end ECG classification using deep neural networks \citep{xu2018towards}, while Strodthoff et al. provided an important benchmark study on deep learning for ECG analysis using PTB-XL dataset \citep{strodthoff2020deep}. These studies demonstrated the potential of deep neural networks to learn clinically useful ECG representations directly from waveform data. However, most task-specific deep learning models require large amounts of accurately labelled ECG data, which is expensive and time-consuming to obtain in clinical practice. Moreover, supervised models may show reduced performance when transferred to external datasets because of differences in sampling frequency, lead configuration, acquisition device, noise level, rhythm prevalence, and patient population \citep{ben2024rawecgnet}. This concern has motivated recent works such as RawECGNet, which explicitly studied deep learning generalization for AF detection from raw ECG signals \citep{ben2024rawecgnet}.

\subsection{ECG Foundation Models and Representation Learning}

Foundation models have recently emerged as a promising direction for ECG analysis. Instead of training a model from scratch for each downstream task, foundation models are pretrained on large-scale ECG datasets and then reused as general-purpose representation learners. This strategy can reduce dependence on task-specific labelled data and may improve generalization across datasets and clinical tasks. Early ECG representation learning studies adapted self-supervised and contrastive learning methods to ECG signals, which includes instance-discrimination objectives inspired by SimCLR \citep{chen2020simple,mehari2022self}. 

Several publicly available ECG foundation models have recently been introduced. HuBERT-ECG adapts the masked-prediction objective of the HuBERT speech model to ECG signals by learning to predict latent cluster assignments from masked ECG segments \citep{coppola2024hubert}. CLEF employs clinically guided contrastive learning to align ECG representations with meaningful clinical information and has been reported to transfer effectively to downstream cardiac tasks \citep{shu2025clef}. ST-MEM extends masked autoencoding to multi-lead ECG by using spatio-temporal patchification and reconstruction-based pretraining \citep{na2024guiding}. ECG-JEPA adapts the joint-embedding predictive architecture to ECG by predicting representations of masked target regions rather than reconstructing the raw signal \citep{kim2024learning}. ECGFounder, in contrast, is pretrained in a supervised manner on a large curated clinical ECG dataset and has shown strong transferability across multiple ECG diagnostic tasks \citep{li2025electrocardiogram}. These models represent different architectural and pretraining choices, including convolutional encoders, transformers, contrastive learning, masked prediction, masked autoencoding, joint-embedding prediction, and large-scale supervised pretraining.

\subsection{Research Gap}

Although ECG foundation models have shown promising results, their relative effectiveness for AF detection remains unclear. Existing studies often evaluate each model using different datasets, preprocessing pipelines, segmentation strategies, downstream classifiers, and validation protocols. As a result, it is difficult to determine whether reported performance differences are caused by the pretrained representations themselves or by inconsistencies in the experimental design. This issue is particularly important because previous studies have shown that deep ECG models can be strongly affected by dataset characteristics, recording duration, lead configuration, and generalization settings \citep{strodthoff2020deep,zhang2023variable,ben2024rawecgnet}.

Computational efficiency is another important but underexplored aspect. For practical AF screening in edge, wearable, and bedside monitoring systems, high classification accuracy alone is not sufficient. Model size, memory usage, inference time, and the need for task-specific fine-tuning are also critical for real-world deployment. Recent studies have emphasized the feasibility of AI-based decision support for large-scale AF screening and the potential implementation of deep neural networks in wearable devices \citep{lueken2025towards}. However, a unified benchmark comparing multiple ECG foundation models under the same frozen-encoder, downstream classification, statistical testing, and computational profiling protocol is still lacking. This study addresses this gap by evaluating nine publicly available ECG foundation models across four heterogeneous ECG datasets using a unified pipeline, recording-level grouped cross-validation, paired recording-level bootstrap comparisons with Holm correction, and computational efficiency profiling.

\section{Materials and Methods}
\label{sec:materialsandmethods}
\subsection{Datasets}

Four publicly available ECG datasets from PhysioNet \citep{goldberger2000physiobank} were used to benchmark the nine foundation models for AF detection. The datasets were selected to cover a wide range of recording conditions so that the generalization behavior of each of the models could be assessed across heterogeneous acquisition settings. A complete profile of the four datasets is given in Table~\ref{tab:datasets}.

The MIT-BIH AF Database (AFDB) \citep{moody1983new} consists of long-term two-channel ambulatory ECG recordings sampled at 250~Hz. In this study, 23 recordings were used, and lead II was selected for analysis. Rhythm annotations were parsed to assign each segment a binary label, retaining only the normal (N) and AF/atrial flutter (AF/AFL) classes. AFDB contributed 168,645 ten-second windows, of which 40.6\% were labelled as AF.

The PhysioNet/Computing in Cardiology Challenge 2017 dataset (CinC2017) \citep{clifford2017af} contains short single-lead ECG recordings sampled at 300~Hz. The original dataset defines four classes, namely N, AF, other rhythm, and noisy recordings. Consistent with the binary AF-detection objective of this study, only the N and AF recordings were retained, yielding 5,777 recordings. CinC2017 is the only single-lead dataset in this study and represents the most challenging acquisition condition owing to its short recording length and low AF prevalence. Segmentation produced 30,825 ten-second windows with an AF prevalence of 12.8\%.

The MIT-BIH Long-Term AF Database (LTAFDB) \citep{petrutiu2007abrupt} comprises 84 long-term, two-channel Holter ECG recordings sampled at 128~Hz from patients with paroxysmal or sustained AF. For this study, Lead II was selected for processing. Due to its extended recording durations, LTAFDB is the largest dataset utilized in this work, contributing 1,411,548 ten-second segments with a well-balanced AF prevalence of 52.6\%.

The China Physiological Signal Challenge 2021 dataset (CPSC2021) \citep{wang2021paroxysmal} consists of variable-length two-lead ECG recordings sampled at 200~Hz and was released for the paroxysmal AF detection challenge. Lead II was used for analysis and 1,424 recordings were processed. Segmentation produced 343,576 ten-second windows with an AF prevalence of 34.3\%.

Across the four datasets, a total of 7,308 recordings were processed, yielding 1,954,594 ten-second ECG segments. A duration of 10~s was selected to provide sufficient temporal context for capturing short-term AF-related rhythm irregularities while maintaining a standardized input length across datasets and models and supporting computationally efficient inference. The specific characteristics of each dataset are summarized in Table~\ref{tab:datasets}. 

\begin{table}[!t]
\caption{Summary of the used ECG datasets.}
\label{tab:datasets}
\scriptsize
\centering
\begin{tabular}{lcccc}
\hline
\textbf{Property} & \textbf{AFDB} & \textbf{CinC2017} & \textbf{LTAFDB} & \textbf{CPSC2021} \\
\hline
Recordings & 23 & 5,777 & 84 & 1,424 \\
Used lead & Lead II & Single lead & Lead II & Lead II \\
Sampling rate & 250 Hz & 300 Hz & 128 Hz & 200 Hz \\
Recording type & Long-term & Short & Long-term & Variable \\
10-s windows & 168,645 & 30,825 & 1,411,548 & 343,576 \\
AF prevalence & 40.6\% & 12.8\% & 52.6\% & 34.3\% \\
\hline
\end{tabular}
\end{table}

\subsection{Preprocessing}

All four datasets were processed through an identical preprocessing pipeline so that any difference in downstream performance could be attributed to the foundation models themselves rather than to inconsistent data preparation. The pipeline consisted of lead selection, filtering, segmentation, labelling, normalisation, and model-specific resampling.

For the two-lead datasets, namely AFDB, LTAFDB, and CPSC2021, lead II was extracted and used for all subsequent analyses, as it is the standard lead for rhythm assessment. For the single-lead CinC2017 dataset, the available channel was used directly. Following lead selection, each continuous ECG recording was filtered before segmentation using a dual-stage filtering procedure. A sixth-order Butterworth band-pass filter with cutoff frequencies of 0.5 and 30~Hz was first applied to attenuate baseline wander and high-frequency noise. A second-order IIR notch filter centred at 50~Hz was subsequently used to suppress residual power-line interference \citep{clifford2006advanced}. Each continuous recording was then divided into fixed-length windows of 10~s using a sliding window with a 5~s stride, which corresponds to 50\% overlap between consecutive windows. The 50\% overlap was selected to reduce sensitivity to arbitrary window boundaries and increase the likelihood that short AF-related rhythm patterns were captured within a complete window, while limiting the redundancy and computational cost associated with smaller strides. Only complete 10~s windows were retained and any trailing segment shorter than 10~s was discarded. In addition, recordings shorter than 10~s were excluded entirely. Because the four datasets were acquired at different sampling rates, a 10~s window corresponded to 1,280, 2,000, 2,500, and 3,000 samples for LTAFDB, CPSC2021, AFDB, and CinC2017, respectively.

\begin{figure*}[htbp]\centering
\begin{subfigure}{0.48\textwidth}
  \centering
  \includegraphics[width=0.95\linewidth]{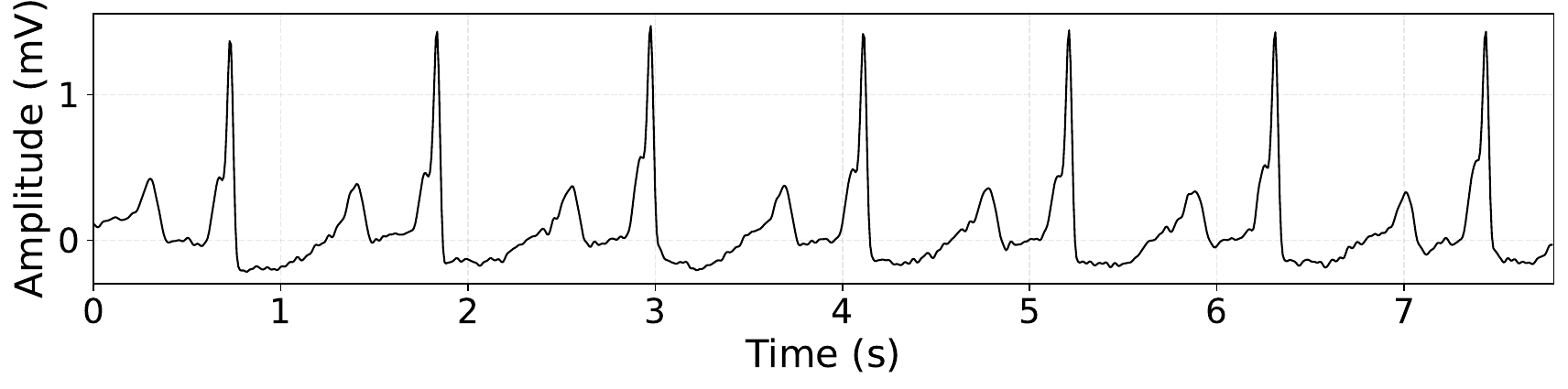}
  \caption{}
  \label{normal}
\end{subfigure}
\hspace{1em}
\begin{subfigure}{0.48\textwidth}
  \centering
  \includegraphics[width=0.95\linewidth]{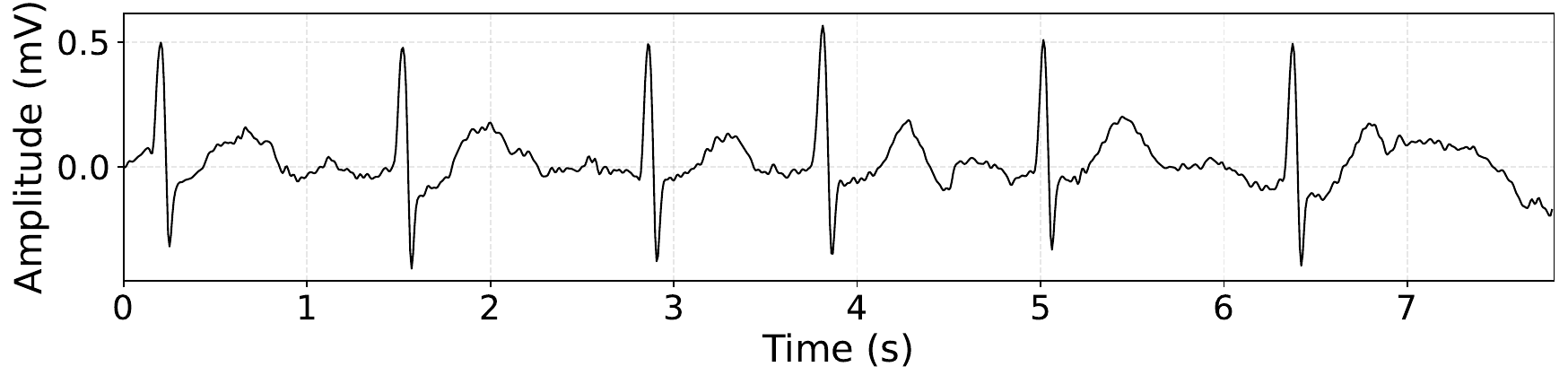}
  \caption{}
  \label{afib}
\end{subfigure}
\caption{An example of (a) Normal and (b) AF segments extracted from the second channel (lead II) of the AFDB dataset. The y-axis represents the ECG signal amplitude in millivolts (mV).}
\label{fig:NormalvsAF}
\end{figure*}

Each window was assigned a binary label by mapping the reference rhythm annotations provided with the datasets to the overlapping 10-second windows generated in this study. The annotations were parsed using the WaveForm Database (WFDB) software package \citep{goldberger2000physiobank,silva2014open}. For AFDB, LTAFDB, and CPSC2021, windows contained within AF or AFL intervals were assigned to the AF class, whereas windows contained within Normal rhythm intervals were assigned to the Normal class. Windows corresponding to other rhythms or noisy annotations were excluded. For CinC2017, each window inherited the original recording-level label. Representative examples of a Normal and an AF window are shown in Fig.~\ref{fig:NormalvsAF}. The Normal rhythm exhibits regular RR intervals and visible P-waves, whereas the AF rhythm shows clear beat-to-beat irregularity and absence of organised P-waves.

Prior to feature extraction, each recording was normalised using z-score normalisation. This step reduced inter-recording amplitude variation and matched the input expectation of the foundation models \citep{strodthoff2020deep}. Any non-finite values were replaced with zero before normalisation to ensure numerical stability. Finally, because each foundation model was pretrained at a specific native sampling rate, every 10~s window was resampled from its dataset-native rate to the native rate of the model under evaluation using a polyphase finite-impulse-response filter \citep{vaidyanathan2006multirate}, which preserves ECG morphology while avoiding aliasing. Performing resampling per model, rather than imposing a single common sampling rate, ensured that each model received input at the resolution for which it was designed.

\subsection{ECG Foundation Models}

Nine publicly released ECG foundation models, from five model families, were evaluated in this study. These models were selected for three main reasons. First, they are openly available with pretrained weights and reproducible loading code. Second, they cover the principal architectural and pretraining strategies currently used for ECG representation learning, including 1-D CNNs, Vision Transformers, masked prediction, contrastive learning, masked autoencoding, joint-embedding predictive learning, and supervised pretraining. Third, they span a parameter range from less than one million to nearly three hundred million parameters, which allows the influence of both architecture and scale on downstream AF detection to be examined under a single protocol. All models were used as frozen feature extractors and their pretrained weights were loaded and kept fixed. Therefore, each model was used only to convert a 10~s ECG window into a fixed-length embedding vector. A summary of the nine models is given in Table~\ref{tab:foundation_models}.

\begin{table*}[!t]
\caption{Summary of the evaluated ECG foundation models.}
\label{tab:foundation_models}
\centering
\vspace{1mm}

\scriptsize
\setlength{\tabcolsep}{2.5pt}
\renewcommand{\arraystretch}{1.18}
\setlength{\extrarowheight}{0.8mm}

\renewcommand{\tabularxcolumn}[1]{m{#1}}

\begin{tabularx}{\textwidth}{
@{}
L{2.4cm}
C{2.9cm}
C{2.30cm}
C{1.35cm}
C{2.05cm}
Y
@{}
}
\hline

\multicolumn{1}{c}{\textbf{Model}} &
\makecell[c]{\textbf{Variants}} &
\makecell[c]{\textbf{Parameters}\\\textbf{(million)}} &
\makecell[c]{\textbf{Native rate}\\\textbf{(Hz)}} &
\makecell[c]{\textbf{Encoder type}} &
\makecell[c]{\textbf{Pretraining dataset(s)}} \\
\hline

HuBERT-ECG \citep{coppola2024hubert} &
\mbox{Small / Base / Large} &
\mbox{30.5 / 93.1 / 188.6} &
100 &
Transformer &
\makecell[c]{11 datasets comprising approximately \\
9.1 million 12-lead ECGs from four countries} \\

CLEF \citep{shu2025clef} &
\mbox{Small / Medium / Large} &
\mbox{0.4 / 30.7 / 296.6} &
500 &
1-D CNN &
MIMIC-IV-ECG \\

ST-MEM \citep{na2024guiding} &
-- &
85.2 &
250 &
Vision Transformer &
Chapman, Ningbo, and CODE-15 \\

ECG-JEPA \citep{kim2024learning} &
-- &
85.4 &
500 &
Vision Transformer &
\makecell[c]{MIMIC-IV-ECG, CODE-15, PTB-XL,\\
Chapman-Shaoxing, CPSC2018, CPSC-Extra,\\
Georgia, Ningbo, PTB, and St-Petersburg} \\

ECGFounder \citep{li2025electrocardiogram} &
-- &
30.7 &
500 &
1-D CNN &
Harvard--Emory ECG Database \\

\hline
\end{tabularx}

\vspace{2pt}

\begin{minipage}{\textwidth}
\footnotesize
\textit{Note:} Based on the pretraining datasets reported in the original
publications, none of the four evaluation datasets used in this study was
explicitly included during model pretraining. For models trained on private
or partially disclosed cohorts, the assessment is limited to the information
available in the original publications.
\end{minipage}

\end{table*}

HuBERT-ECG \citep{coppola2024hubert} adapts the masked-prediction objective of the speech model HuBERT \citep{hsu2021hubert} to the ECG domain by learning representations through the prediction of cluster assignments from masked signal segments. Three transformer variants, namely Small, Base, and Large, were evaluated. CLEF \citep{shu2025clef} is a convolutional ECG foundation model trained with a multi-scale contrastive objective, and its Small, Medium, and Large variants were evaluated. ST-MEM \citep{na2024guiding} is a Vision Transformer encoder pretrained with a spatio-temporal masked-autoencoding objective, extending the masked-autoencoder paradigm \citep{he2022masked} to multi-lead ECG. ECG-JEPA \citep{kim2024learning} is a Vision Transformer encoder pretrained with a joint-embedding predictive objective \citep{assran2023self}, in which the representation of a masked target region is predicted rather than the raw signal. ECGFounder \citep{li2025electrocardiogram} is a one-dimensional convolutional encoder pretrained in a supervised manner on a large-scale curated clinical ECG corpus.

For every model, each 10~s window was passed through the frozen encoder, and the resulting hidden representation was reduced to a single embedding vector per window by global pooling when the encoder produced a sequence of tokens. These embeddings were then used as input features for the downstream classifier described in Section~\ref{subsec:downstream_classification}.

\subsection{Downstream Classification}
\label{subsec:downstream_classification}

The frozen embeddings produced by each foundation model were classified using the XGBoost algorithm \citep{chen2016xgboost}, a scalable implementation of gradient boosting \citep{friedman2001greedy}. XGBoost was selected because it can efficiently handle high-dimensional and heterogeneous feature representations, capture nonlinear interactions between embedding dimensions, and produce deterministic results under a fixed random seed. A separate classifier was trained for each model--dataset combination so that the reported performances reflect the quality of each model's embeddings rather than the capacity of the classifier.

The XGBoost classifier was configured with 300 estimators, a maximum tree depth of 6, a learning rate of 0.1, a sub-sampling ratio of 0.8, and a column-subsampling ratio of 0.8. Although the dimensionality of the extracted embeddings may differ across foundation models, XGBoost can operate on feature vectors of varying dimensions. A separate classifier was trained for each model and dataset combination, while the same hyperparameters were retained to ensure a standardized downstream learning setting and avoid model-specific tuning as a confounding factor. To address class imbalance, the positive-class weight was set automatically for each training fold according to the ratio of Normal to AF windows in that fold, and a fixed decision threshold of 0.5 was applied. No exhaustive hyperparameter tuning was performed and the configuration was fixed across all models and datasets to ensure a fair and reproducible comparison. Within each cross-validation fold, the embedding features were standardized using z-score scaling, with the scaler fitted only on the training set and then applied to the test/validation set. This ensured that no information from the train data leaked to the testing phase.

Model performance was estimated using 5-fold cross-validation with a grouped splitting strategy in which grouping was performed by recording identity \citep{saeb2017need}. All windows originating from a given recording were assigned to the same fold, preventing temporally correlated windows from the same recording from appearing in both the training and test/validation folds. This strategy ensures avoiding the optimistic bias that a naive window-level split could introduce. For each of the five folds, the classifier was trained on four folds and evaluated on the held-out fold. The reported metrics are the mean and standard deviation computed across the five folds.

\section{The FOUND-AF Unified Benchmarking Framework}
\label{sec:pipeline}

\begin{figure}[!htpb]
    \centering
    \includegraphics[width=0.99\linewidth]{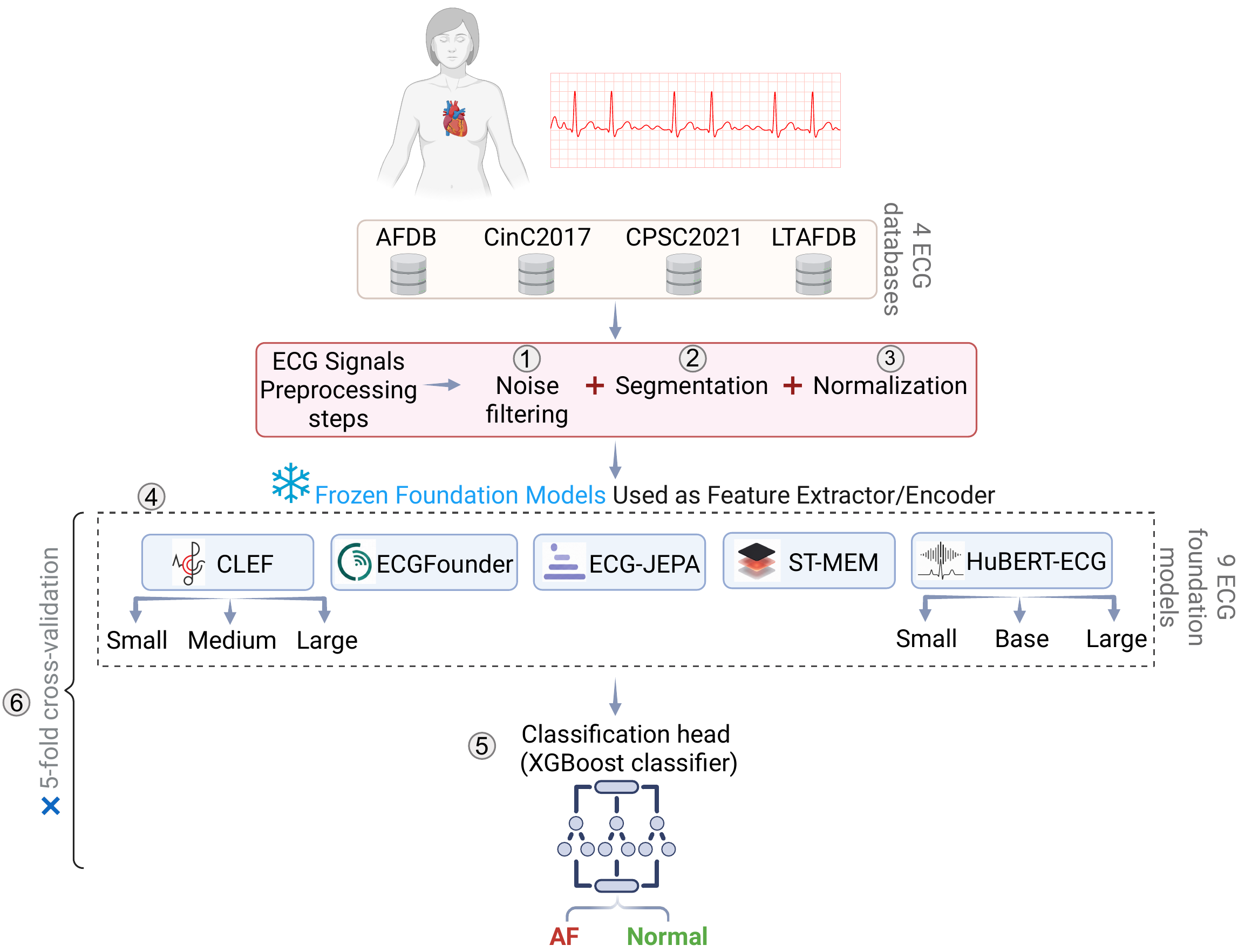}
    \caption{Flowchart of the proposed FOUND-AF evaluation pipeline. Four ECG datasets are first preprocessed and segmented into fixed-length windows. The resulting ECG windows are then passed through frozen ECG foundation models to extract embeddings, which are subsequently classified using an XGBoost classifier under a recording-level 5-fold cross-validation protocol.}
    \label{fig:flowchart}
\end{figure}

Figure~\ref{fig:flowchart} shows the overall experimental flowchart used to benchmark the ECG foundation models for AF detection. The pipeline was designed to ensure a fair comparison across all evaluated models by using the same datasets, preprocessing steps, validation strategy, downstream classifier, and performance metrics.

The main steps of the pipeline are summarized as follows:

\begin{enumerate}

    \item \textit{Signal filtering}: The selected ECG channel is filtered using a sixth-order Butterworth band-pass filter with cutoff frequencies of 0.5--30~Hz to attenuate baseline wander and high-frequency noise. A second-order IIR notch filter centred at 50~Hz is subsequently applied to suppress power-line interference.

    \item \textit{Segmentation and labeling}: Each ECG recording is segmented into fixed-length 10~s windows using a 5~s stride, which corresponds to 50\% overlap between consecutive windows. Each window is assigned a binary label, AF or Normal, according to the reference rhythm annotations.

    \item \textit{Normalization}: Each ECG segment is normalized to zero mean and unit standard deviation to mitigate inter-segment amplitude variations and improve numerical stability during feature embedding/extraction and classification.

    \item \textit{Frozen foundation model encoding}: The preprocessed ECG windows are passed through the ECG foundation models, which are used as frozen feature extractors. The pretrained weights are not updated during this process. Nine models from five model families are evaluated.

    \item \textit{Classifier training}: The embeddings extracted from the foundation models are used to train an XGBoost classifier. A separate classifier is trained for each model--dataset combination to ensure that the reported performance reflects the discriminative quality of each foundation model embedding.

    \item \textit{Model evaluation}: The trained classifier is then evaluated using 5-fold cross-validation strategy. Performance is assessed using standard classification metrics, including accuracy, sensitivity, specificity, precision, F\textsubscript{1}-score, and ROC-AUC. The final results are reported as the mean and standard deviation across the five folds.

\end{enumerate}

This pipeline is repeated independently for all foundation models and datasets. By holding the preprocessing, downstream classifier, validation protocol, and decision threshold fixed, FOUND-AF minimizes experimental confounding and enables observed performance differences to be attributed primarily to the quality of the pretrained ECG representations.

\section{Experimental Validation}
\label{sec:results}

This section reports the experimental evaluation of nine ECG foundation models across four datasets, resulting in 36 distinct model–dataset combinations. The following subsections present the evaluation metrics, the complete classification results, statistical comparisons, a focused analysis of the best-performing model, and an assessment of inference time and memory usage.

\subsection{Evaluation Metrics}
\label{subsec:evaluationmetrics}

To assess the performance of the AF detection models, five standard classification metrics are utilized. The classification metrics used in this study are accuracy (\emph{Acc}), sensitivity (\emph{Se}), specificity (\emph{Sp}), precision (\emph{Pr}) and \emph{F}-score as given below.

\begin{equation}
    Acc = \frac{TP + TN}{TP + FN + FP + TN},
\label{eq:accuracy}
\end{equation}

\begin{equation}
    Se = \frac{TP}{TP + FN},
\label{eq:sensitivity}
\end{equation}

\begin{equation}
    Sp = \frac{TN}{TN + FP},
\label{eq:specificity}
\end{equation}

\begin{equation}
    Pr = \frac{TP}{TP + FP},
\label{eq:precision}
\end{equation}

\begin{equation}
    F-score = (1 + \beta)\frac{Pr \cdot Se}{(\beta^2 \cdot Pr) + Se},
\label{eq:f1}
\end{equation}
where \emph{TP}, \emph{TN}, \emph{FP}, and \emph{FN} denote the number of true positives, true negatives, false positives, and false negatives, respectively. The \emph{F}-score represents the harmonic mean of sensitivity (\emph{Se}) and precision (\emph{Pr}). When $\beta = 1$, it is called the balanced \emph{F}\textsubscript{1}-score, which assigns equal importance to sensitivity and precision in its computation and is used in this study \citep{powers2020evaluation}.

\subsection{Comparison of ECG foundation models}
\label{subsec:compFoundation}

Table~\ref{tab:classification_performance} summarizes the classification performance of the nine ECG foundation models across the four datasets. For each dataset and each metric, the best value is highlighted in bold, and the second-best value is underlined. Across all 36 experiments, ECGFounder achieved the strongest performance, ranking first on every dataset for accuracy, sensitivity, precision, and F\textsubscript{1}-score, with F\textsubscript{1}-scores of 97.87\% on AFDB, 92.77\% on CinC2017, 99.50\% on CPSC2021, and 95.68\% on LTAFDB. The transformer-based self-supervised models ECG-JEPA and ST-MEM consistently formed the second tier, while the HuBERT-ECG family and the smaller CLEF variants trailed. 

\begin{table*}[!t]
\caption{Classification performance of ECG foundation models across four datasets. The best model for each evaluation metric within each dataset is highlighted in bold and the second-best value is underlined. All values are in percentage (\%).}
\label{tab:classification_performance}
\centering
\scriptsize
\setlength{\tabcolsep}{1.2pt}
\renewcommand{\arraystretch}{1.15}
\resizebox{\textwidth}{!}{%
\begin{tabular}{@{\extracolsep{8pt}}lccccc ccccc ccccc ccccc}
\hline
\textbf{Model} 
& \multicolumn{5}{c}{\textbf{AFDB}} 
& \multicolumn{5}{c}{\textbf{CinC2017}} 
& \multicolumn{5}{c}{\textbf{CPSC2021}} 
& \multicolumn{5}{c}{\textbf{LTAFDB}} \\
\cline{2-6} \cline{7-11} \cline{12-16} \cline{17-21}
& Acc & Se & Pr & Sp & F\textsubscript{1}-score
& Acc & Se & Pr & Sp & F\textsubscript{1}-score
& Acc & Se & Pr & Sp & F\textsubscript{1}-score
& Acc & Se & Pr & Sp & F\textsubscript{1}-score \\
\hline
ECGFounder 
& \textbf{98.33} & \textbf{97.04} & \textbf{98.73} & \textbf{99.11} & \textbf{97.87}
& \textbf{98.15} & \textbf{92.86} & \textbf{92.70} & \textbf{98.92} & \textbf{92.77}
& \textbf{99.66} & \textbf{99.65} & \textbf{99.36} & \textbf{99.67} & \textbf{99.50}
& \textbf{95.49} & \textbf{96.81} & \textbf{94.67} & \textbf{94.27} & \textbf{95.68} \\

ECG-JEPA 
& \underline{93.67} & \underline{91.03} & \underline{92.97} & 95.57 & \underline{91.46}
& 97.33 & 87.81 & 91.03 & 98.72 & 89.38
& \underline{99.45} & \underline{99.22} & \underline{99.16} & \underline{99.57} & \underline{99.19}
& \underline{93.11} & \underline{93.78} & \underline{92.86} & \underline{92.86} & \underline{93.23} \\

ST-MEM 
& 92.63 & 87.74 & 92.38 & \underline{95.74} & 89.59
& \underline{97.47} & \underline{88.18} & \underline{91.64} & \underline{98.82} & \underline{89.87}
& 99.04 & 98.81 & 98.38 & 99.15 & 98.59
& 91.61 & 91.28 & 92.46 & 92.08 & 91.69 \\

CLEF-Small 
& 74.77 & 69.86 & 71.76 & 79.66 & 67.38
& 93.79 & 69.64 & 79.08 & 97.31 & 74.04
& 93.67 & 90.89 & 90.74 & 95.15 & 90.77
& 81.79 & 83.63 & 81.50 & 79.74 & 82.40 \\

CLEF-Medium 
& 86.04 & 82.33 & 83.59 & 88.67 & 81.88
& 95.83 & 79.64 & 86.68 & 98.19 & 82.96
& 95.38 & 94.80 & 91.96 & 95.69 & 93.34
& 84.14 & 84.37 & 85.26 & 84.21 & 84.51 \\

CLEF-Large 
& 83.56 & 85.00 & 79.41 & 81.79 & 79.77
& 93.51 & 69.94 & 77.03 & 96.94 & 73.28
& 94.98 & 95.75 & 90.17 & 94.57 & 92.87
& 87.01 & 88.83 & 86.62 & 85.44 & 87.49 \\

HuBERT-ECG-Small 
& 70.93 & 60.00 & 68.87 & 81.38 & 62.12
& 92.32 & 60.89 & 74.17 & 96.90 & 66.82
& 95.79 & 94.35 & 93.39 & 96.54 & 93.86
& 82.35 & 83.62 & 82.28 & 80.90 & 82.74 \\

HuBERT-ECG-Base 
& 81.06 & 77.75 & 75.46 & 84.15 & 75.85
& 93.75 & 65.06 & 82.22 & 97.94 & 72.63
& 94.67 & 93.39 & 91.17 & 95.31 & 92.26
& 82.37 & 83.12 & 82.96 & 81.79 & 82.81 \\

HuBERT-ECG-Large 
& 76.99 & 66.14 & 72.18 & 84.75 & 67.76
& 91.71 & 57.42 & 71.87 & 96.72 & 63.83
& 91.94 & 89.53 & 87.18 & 93.17 & 88.33
& 79.52 & 81.29 & 79.51 & 77.40 & 80.16 \\
\midrule
\end{tabular}%
}
\end{table*}

The results in Table~\ref{tab:classification_performance} reveal a consistent ranking of model families together with clear dataset-specific patterns. On the CPSC2021 dataset, all three leading models performed exceptionally well, with ECGFounder, ECG-JEPA, and ST-MEM reaching F\textsubscript{1}-scores of 99.50\%, 99.19\%, and 98.59\%, respectively. Even the weakest model on this dataset, HuBERT-ECG-Large, achieved an F\textsubscript{1}-score of 88.33\%. Therefore, CPSC2021 represents the dataset on which the evaluated foundation models were most effective overall.

On the LTAFDB dataset, ECGFounder achieved the highest F\textsubscript{1}-score of 95.68\%, followed by ECG-JEPA with 93.23\% and ST-MEM with 91.69\%. The CLEF family showed moderate performance, with CLEF-Large reaching an F\textsubscript{1}-score of 87.49\%, whereas the HuBERT-ECG variants clustered around 80--83\% F\textsubscript{1}-score. On AFDB, ECGFounder again achieved the best performance with an F\textsubscript{1}-score of 97.87\%. However, the gap to the second tier was larger. ECG-JEPA and ST-MEM obtained F1-scores of 91.46\% and 89.59\%, respectively, while the remaining six models remained below 82\%. In particular, HuBERT-ECG-Small and HuBERT-ECG-Large achieved F\textsubscript{1}-scores of 62.12\% and 67.76\%, respectively.

CinC2017 represented the most difficult acquisition condition. On this dataset, ECGFounder achieved the highest F\textsubscript{1}-score of 92.77\%, while ST-MEM achieved the second-best performance with an F\textsubscript{1}-score of 89.87\%, marginally outperforming ECG-JEPA, which achieved 89.38\%. This was the only dataset on which ST-MEM outperformed ECG-JEPA. Sensitivity was the limiting metric on CinC2017. Several models exceeded 96\% specificity but achieved sensitivity below 70\%, which indicates a tendency to miss AF windows in short single-lead recordings. Across all four datasets, the ordering of the model families was relatively stable, with ECGFounder ranking first, the transformer-based self-supervised models ECG-JEPA and ST-MEM forming the second tier, the CLEF family ranking third, and the HuBERT-ECG family showing the lowest overall performance.

\begin{figure*}[!h]
    \centering
    \includegraphics[width=0.95\linewidth]{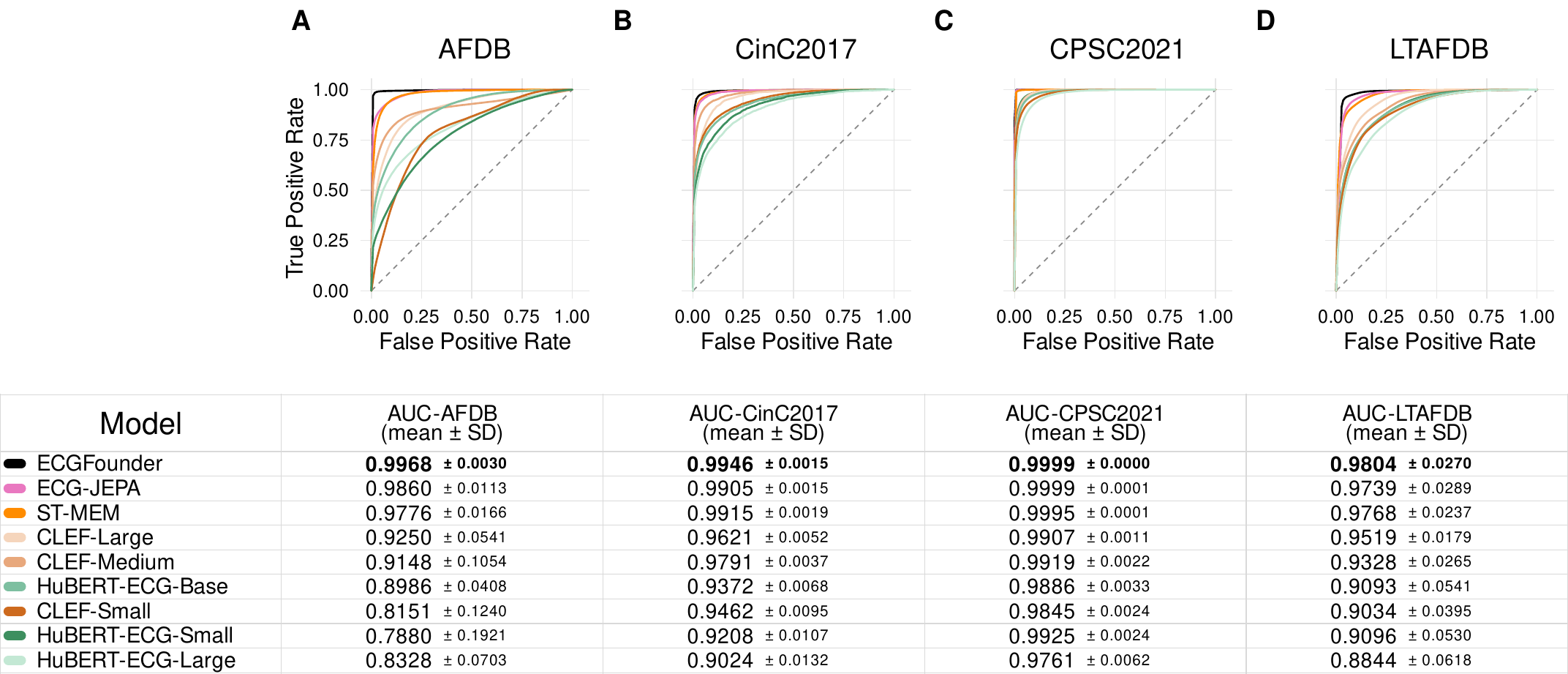}
    \caption{Receiver operating characteristic (ROC) analysis of the nine ECG foundation models across the four datasets. Figures A--D show the ROC curves for AFDB, CinC2017, CPSC2021, and LTAFDB, respectively. The table below the curves reports the corresponding area under the curve (AUC) values as mean~$\pm$~standard deviation across the five cross-validation folds. ECGFounder achieved the highest AUC on all datasets, followed by ECG-JEPA and ST-MEM.}
    \label{fig:roc_curves}
\end{figure*}

Fig.~\ref{fig:roc_curves} presents the receiver operating characteristic (ROC) \citep{fawcett2006introduction} curves of the nine ECG foundation models across the four datasets, together with the corresponding mean AUC values and standard deviations. Overall, the ROC analysis confirms the ranking observed from the threshold-dependent metrics in Table~\ref{tab:classification_performance}. ECGFounder achieved the highest AUC on all four datasets, with values of 0.9968~$\pm$~0.0030 on AFDB, 0.9946~$\pm$~0.0015 on CinC2017, 0.9999~$\pm$~0.0000 on CPSC2021, and 0.9804~$\pm$~0.0270 on LTAFDB. Its ROC curves were consistently closest to the top-left corner, indicating superior discrimination between AF and Normal windows across heterogeneous acquisition conditions.

ECG-JEPA and ST-MEM generally formed the second-best group, although their relative ranking varied across datasets. ECG-JEPA achieved the second-highest AUC on AFDB, CPSC2021, and LTAFDB, whereas ST-MEM slightly outperformed ECG-JEPA on CinC2017. The CPSC2021 dataset showed near-saturated discrimination for the leading models, with ECGFounder, ECG-JEPA, and ST-MEM all reaching AUC values close to 1.00. In contrast, LTAFDB showed larger standard deviations, suggesting greater fold-to-fold variability, likely due to the long-term nature of the recordings and inter-recording heterogeneity. The HuBERT-ECG variants produced the lowest AUC values overall, particularly on AFDB and CinC2017, which is consistent with their lower sensitivity and F\textsubscript{1}-scores in Table~\ref{tab:classification_performance}. These results indicate that ECGFounder provides not only the best fixed-threshold classification performance but also the strongest threshold-independent discriminative ability.

To determine whether the observed performance differences were statistically significant, a paired recording-level bootstrap analysis \citep{efron1994introduction} was performed. ECGFounder, which achieved the best overall performance, was used as the reference model. For each comparison, recordings were sampled with replacement over 10,000 bootstrap iterations, while all windows belonging to each selected recording were retained. The difference in F\textsubscript{1}-score between ECGFounder and each competing model was calculated for every resample, and the corresponding 95\% confidence interval was obtained from the resulting bootstrap distribution.

\begin{figure*}[!h]
    \centering
    \includegraphics[width=0.95\linewidth]{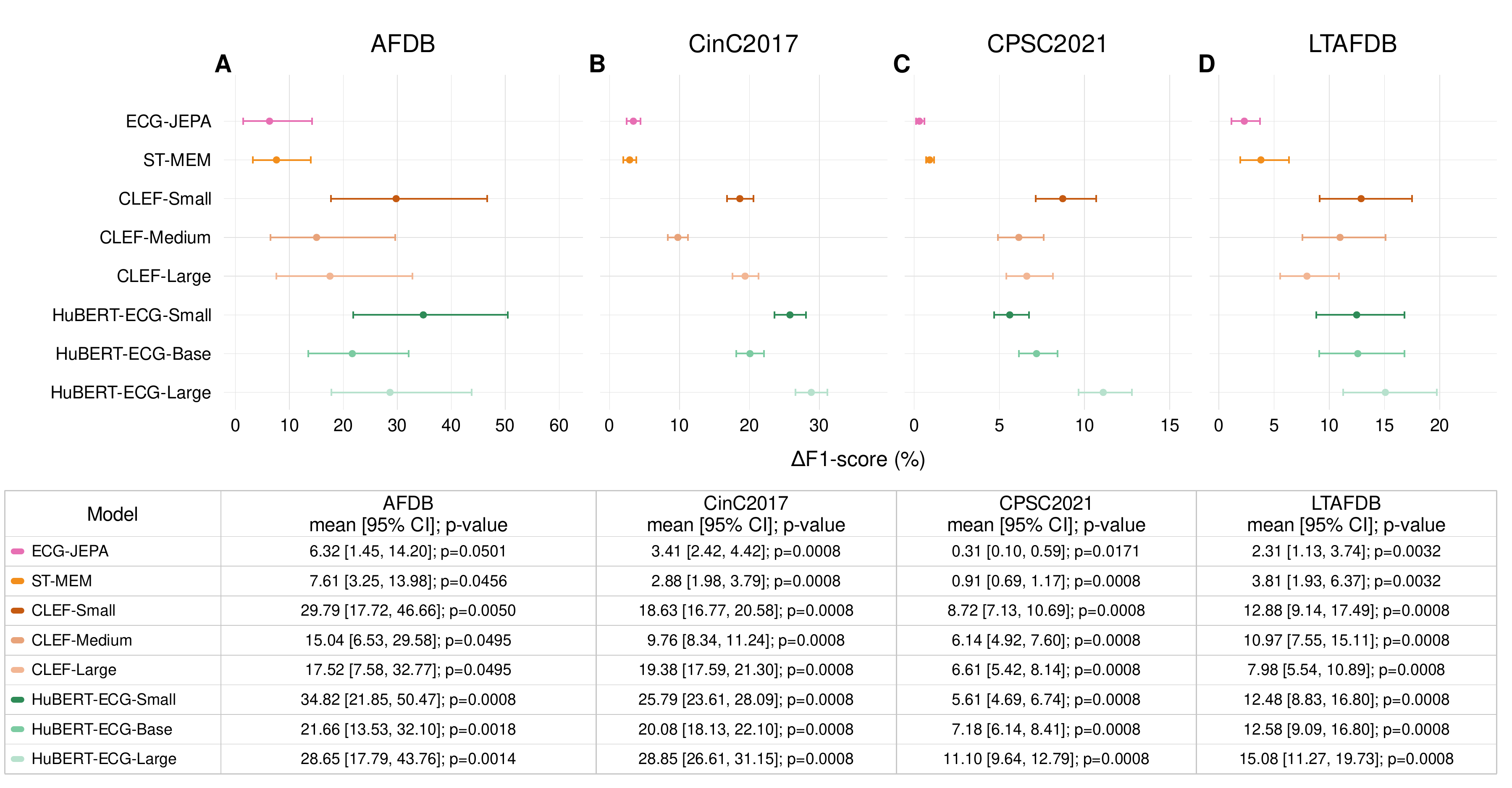}
    \caption{Paired recording-level bootstrap comparison of F\textsubscript{1}-score differences between ECGFounder and the competing ECG foundation models across the four datasets. Figures A--D show the results for AFDB, CinC2017, CPSC2021, and LTAFDB, respectively. Points indicate the F\textsubscript{1}-score difference between ECGFounder and each competing model, and horizontal bars represent the corresponding 95\% confidence intervals obtained from 10,000 recording-level bootstrap resamples. The table reports the difference, confidence interval, and Holm-adjusted $p$-value for each comparison. Positive values favour ECGFounder.}
    \label{fig:bootstrap_forest}
\end{figure*}

Two-sided bootstrap p-values were calculated and adjusted within each dataset using the Holm procedure to account for multiple comparisons. An adjusted $p$-value below 0.05 was considered statistically significant. Fig.~\ref{fig:bootstrap_forest} shows the resulting forest plot. ECGFounder showed positive F\textsubscript{1}-score differences across all comparisons, with statistically significant differences in all comparisons except ECGFounder versus ECG-JEPA on AFDB.

Because ECGFounder achieved the highest F\textsubscript{1}-score on every dataset and showed statistically significant improvements in nearly all pairwise comparisons, the remaining analyses focus primarily on ECGFounder.

\subsection{Performance Analysis of ECGFounder Model}
\label{subsec:ecgfounder}

\begin{figure*}[!htpb]
    \centering
    \includegraphics[width=0.95\linewidth]{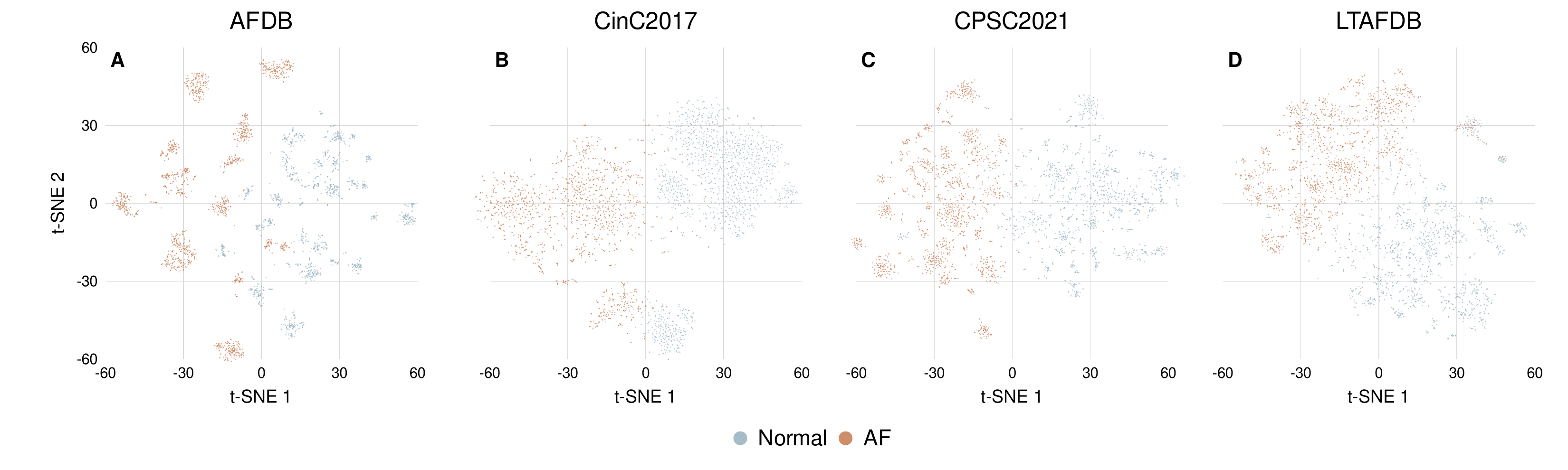}
    \caption{t-SNE visualization of ECGFounder embeddings across the four datasets. Figures A--D show the two-dimensional projections for AFDB, CinC2017, CPSC2021, and LTAFDB, respectively. Each point represents a 10~s ECG window, with colors indicating Normal and AF classes. The embeddings show clear class-dependent structure across datasets, supporting the discriminative capability of ECGFounder as a frozen feature extractor.}
    \label{fig:ecgfounder_tsne}
\end{figure*}

Based on the comparative results presented in Section~\ref{subsec:compFoundation}, ECGFounder was selected for further analysis as the best-performing ECG foundation model. To provide additional insight beyond the aggregate classification metrics, we further examined the structure of the learned embedding space generated by ECGFounder. Specifically, we investigated whether the frozen representations preserve class-discriminative information between AF and Normal rhythm windows across the four datasets. For this purpose, the high-dimensional ECGFounder embeddings were projected into a two-dimensional space using t-distributed stochastic neighbor embedding (t-SNE) \citep{van2008visualizing}, as shown in Fig.~\ref{fig:ecgfounder_tsne}. This analysis was used to visually assess feature separability and to better understand whether the superior downstream classification performance of ECGFounder is supported by a meaningful organization of the extracted representations.

As shown in Fig.~\ref{fig:ecgfounder_tsne}, the ECGFounder embeddings exhibit clear class-dependent structure across all datasets. In AFDB and CPSC2021, AF and Normal windows form largely separated regions, which is consistent with the high F\textsubscript{1}-scores and AUC values reported for these datasets. CinC2017 also shows visible separation between the two rhythm classes, although with a higher degree of overlap, which reflects the more challenging nature of short single-lead recordings. In LTAFDB, the embeddings show broader and more heterogeneous clusters, likely due to the long-term Holter recordings and greater inter-recording variability, nevertheless, AF and Normal samples remain distinguishable in the projected space. Overall, the t-SNE analysis supports the quantitative results by showing that ECGFounder generates discriminative representations that preserve clinically meaningful rhythm differences across heterogeneous ECG datasets.

\begin{figure*}[!htpb]
    \centering
    \includegraphics[width=0.95\linewidth]{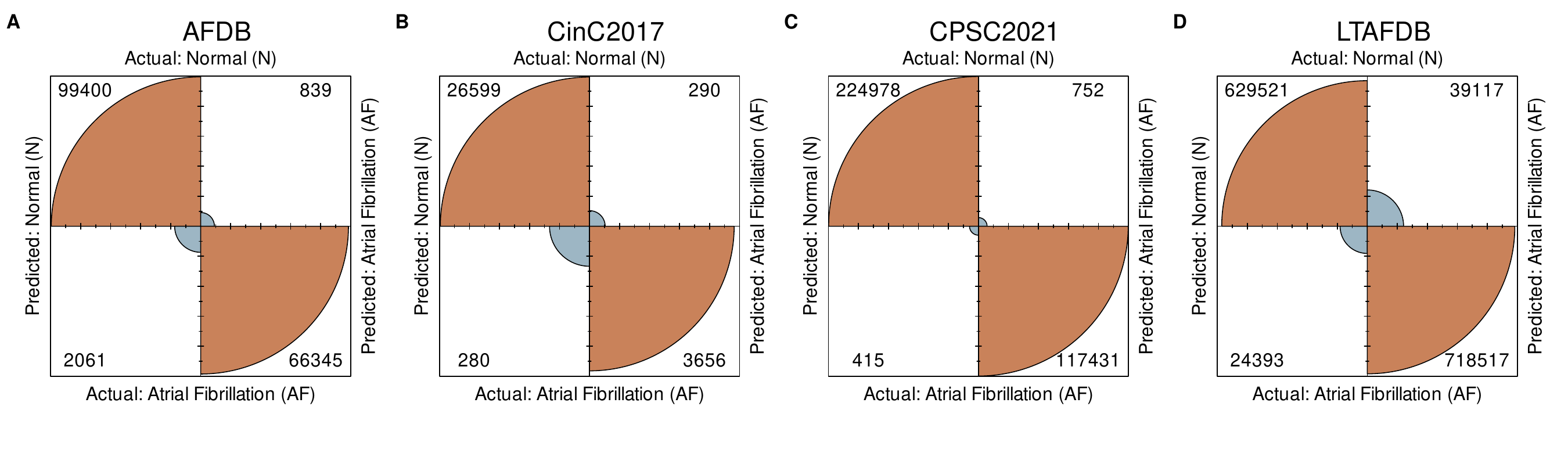}
    \caption{Fourfold confusion-matrix plots for ECGFounder across the four datasets. Figures A--D show the classification results for AFDB, CinC2017, CPSC2021, and LTAFDB, respectively. Each panel summarizes the number of true Normal, true AF, false positive, and false negative windows obtained under the recording-level cross-validation protocol. ECGFounder shows consistently strong discrimination between Normal and AF windows across all datasets.}
    \label{fig:ecgfounder_fourfold}
\end{figure*}

Figure~\ref{fig:ecgfounder_fourfold} presents the fourfold confusion-matrix plots for ECGFounder across the four datasets. Overall, the results confirm the strong classification performance observed in the quantitative evaluation. On AFDB, ECGFounder correctly classified 99,400 Normal windows and 66,345 AF windows, with relatively few false positives and false negatives. Similarly, on CinC2017, the model maintained a balanced classification behaviour, with 26,599 correctly identified Normal windows and 3,656 correctly identified AF windows. Although CinC2017 is the most challenging dataset due to its short single-lead recordings and lower AF prevalence, the number of misclassified windows remained limited.

The best classification behaviour was observed on CPSC2021, where ECGFounder produced very few errors relative to the number of correctly classified windows, with 224,978 true Normal predictions and 117,431 true AF predictions. This is consistent with the near-perfect F\textsubscript{1}-score and AUC obtained on this dataset (Fig.~\ref{fig:roc_curves}). For LTAFDB, the model correctly classified a large number of both Normal and AF windows, with 629,521 true Normal predictions and 718,517 true AF predictions. However, the absolute number of errors was also higher than in the other datasets, which is expected given the substantially larger number of windows and the heterogeneous long-term Holter recordings. Taken together, the fourfold plots show that ECGFounder achieved strong and balanced discrimination between AF and Normal rhythms across different acquisition settings, while maintaining a relatively low number of false predictions.

\subsection{Inference Time, Memory Usage, and Implications for Edge Deployment}
\label{sebsec:edge}

In addition to classification performance, the computational efficiency of each ECG foundation model was evaluated to assess its suitability for edge-based AF detection. All models were profiled under identical conditions using a GPU-enabled Google Colab environment. The reported measurements therefore provide a standardized relative comparison across models, although the absolute values may differ on dedicated edge hardware. Two complementary efficiency indicators were considered: peak resident set size (RSS) memory and inference time per ECG window. Peak RSS memory refers to the maximum amount of physical memory occupied by the process during embedding extraction \citep{siebenmann2012understanding}. It includes the memory required to load the model, store intermediate activations, and process the input window. Therefore, RSS memory provides a practical estimate of the runtime memory footprint of each model, which is more informative for deployment than the parameter count alone. Inference time was defined as the average time required by the frozen encoder to process one 10~s ECG window and generate its embedding. This measure reflects the latency of the feature extraction stage, which is the computationally dominant component of the proposed pipeline before the lightweight XGBoost classifier is applied.

\begin{figure*}[!htbp]\centering
\begin{subfigure}{0.48\textwidth}
  \centering
  \includegraphics[width=0.99\linewidth]{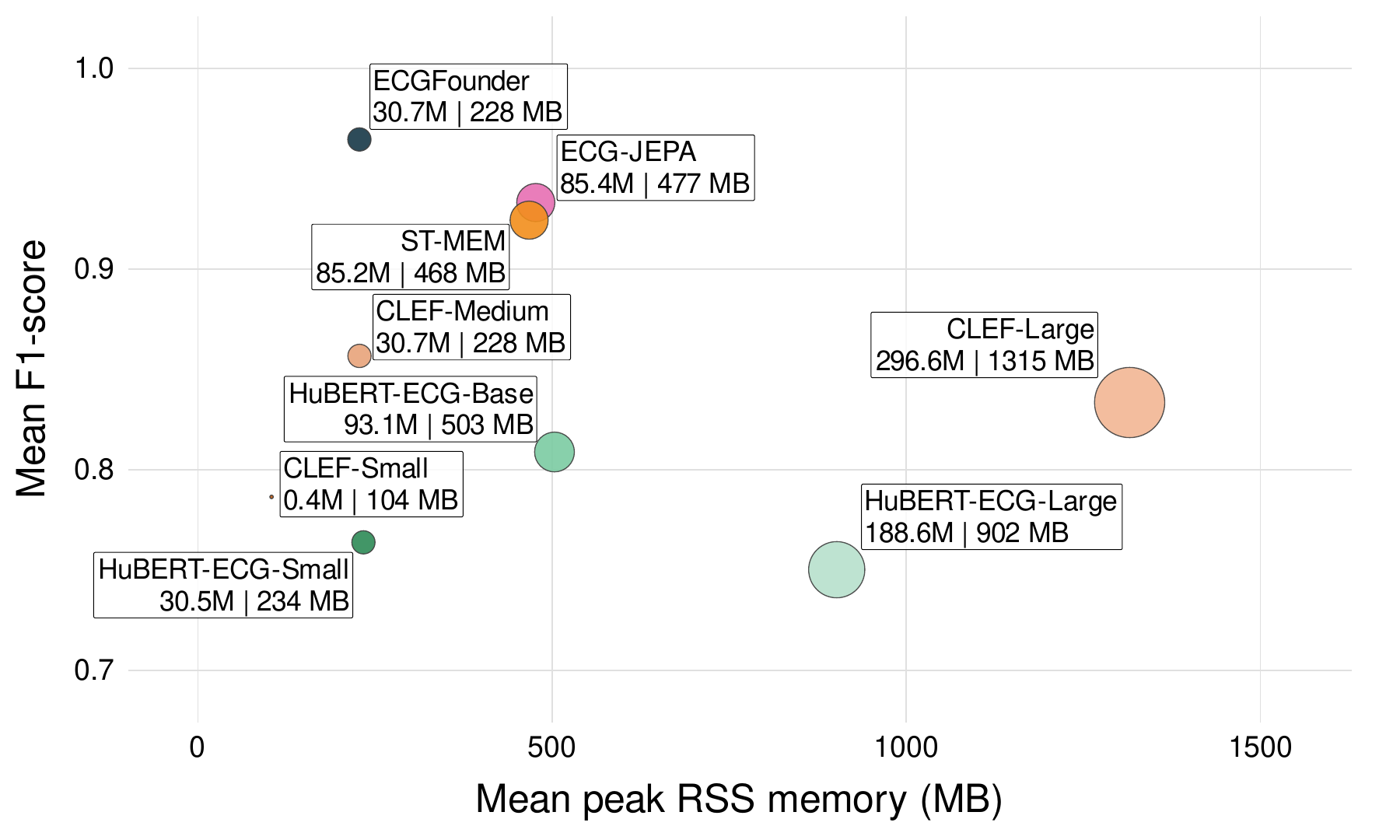}
  \caption{}
  \label{fig:memory_vs_f1}
\end{subfigure}
\begin{subfigure}{0.48\textwidth}
  \centering
  \includegraphics[width=0.99\linewidth]{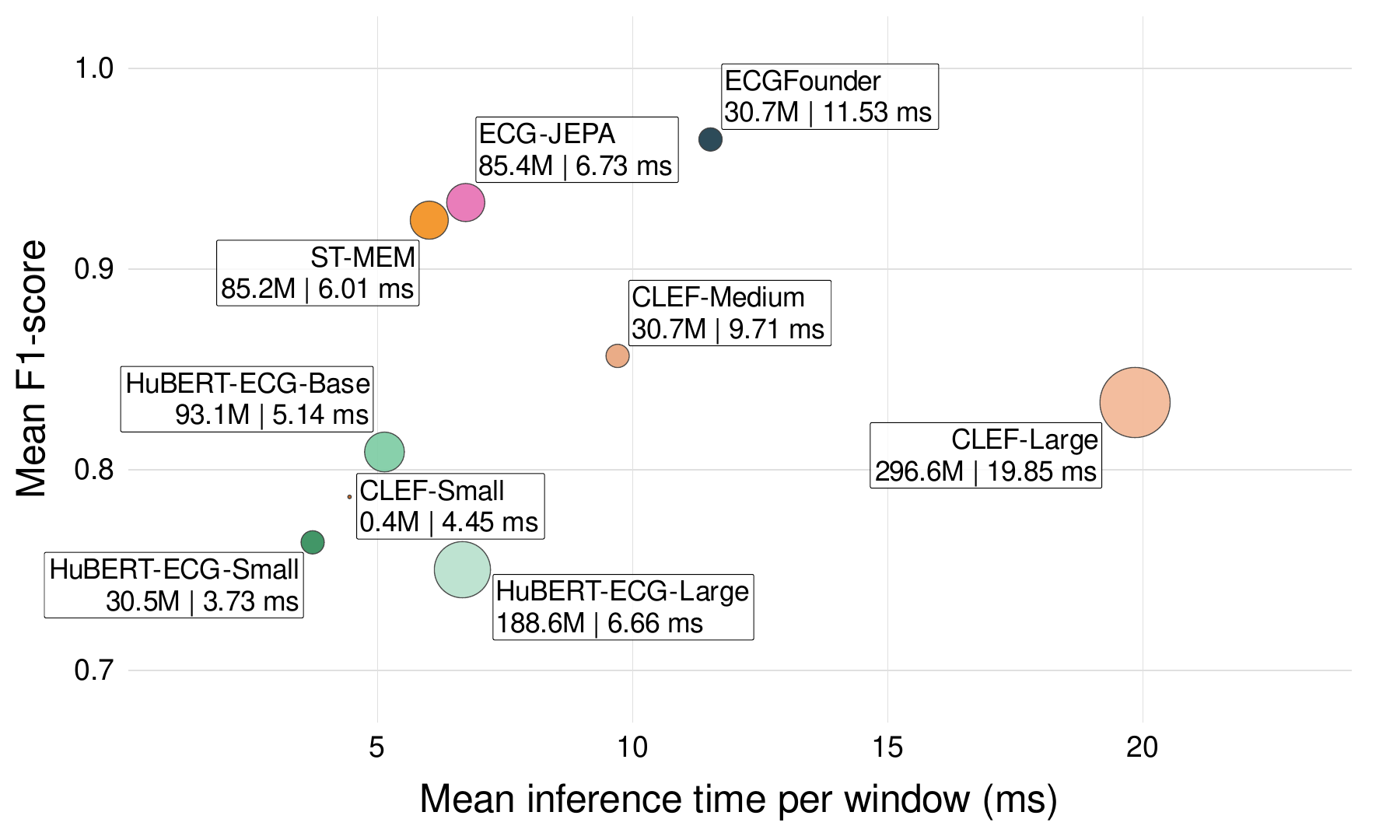}
  \caption{}
  \label{fig:time_vs_f1}
\end{subfigure}
\caption{Accuracy--efficiency trade-off of the evaluated ECG foundation models. 
(a) Relationship between mean F\textsubscript{1}-score and mean peak resident set size (RSS) memory. 
(b) Relationship between mean F\textsubscript{1}-score and mean inference time per 10~s ECG window. 
Each circle represents one foundation model, and the circle size is proportional to the number of model parameters. The text labels report the parameter count together with the corresponding memory usage in (a) or inference time in (b). ECGFounder achieved the highest mean F\textsubscript{1}-score while maintaining moderate memory usage and acceptable inference time, indicating the most favorable trade-off for edge-based AF detection.}
\label{fig:efficiency_tradeoff}
\end{figure*}

Figure~\ref{fig:efficiency_tradeoff} shows the relationship between mean F\textsubscript{1}-score and computational cost. Fig.~\ref{fig:memory_vs_f1} compares mean F\textsubscript{1}-score against mean peak RSS memory, whereas Fig.~\ref{fig:time_vs_f1} compares mean F\textsubscript{1}-score against mean inference time per 10~s window. In both figures, the size of each circle is proportional to the number of model parameters, which allows the relationship between model scale, predictive performance, and deployment cost to be visualized simultaneously.

As shown in Fig.~\ref{fig:memory_vs_f1}, ECGFounder achieved the most favorable accuracy--memory trade-off. It obtained the highest mean F\textsubscript{1}-score while requiring only 228~MB of peak RSS memory and containing 30.7 million parameters. This memory footprint is moderate compared with larger models such as CLEF-Large and HuBERT-ECG-Large, which required 1,315~MB and 902~MB of peak RSS memory, respectively, while achieving substantially lower mean F\textsubscript{1}-scores. ECG-JEPA and ST-MEM also achieved strong classification performance, but required approximately twice the memory of ECGFounder, with peak RSS memory values of 477~MB and 468~MB, respectively. In contrast, CLEF-Small had the lowest memory requirement at 104~MB, but its lower F\textsubscript{1}-score indicates that very compact models may not provide sufficiently discriminative embeddings for robust AF detection.

A similar pattern was observed for inference time in Fig.~\ref{fig:time_vs_f1}. ECGFounder required 11.53~ms per 10~s window, which is compatible with real-time AF screening because the processing latency is several orders of magnitude shorter than the duration of the analyzed ECG window. Although ST-MEM and ECG-JEPA were faster, with inference times of 6.01~ms and 6.73~ms, respectively, they achieved lower mean F\textsubscript{1}-scores than ECGFounder. The fastest models, including HuBERT-ECG-Small and CLEF-Small, required less than 5~ms per window but showed weaker classification performance. Conversely, CLEF-Large had both the largest parameter count and the highest inference time, requiring 19.85~ms per window without achieving competitive performance compared with the leading models.

There is no universal threshold that defines whether a model is feasible for edge AI, as feasibility depends on the target device, available memory, battery capacity, operating system overhead, sampling strategy, and whether inference is performed continuously or intermittently \citep{sze2017efficient}. Nevertheless, for ECG-based AF screening with 10~s windows, a practical model should ideally satisfy two conditions: (1) the inference time should be much shorter than the window duration, and (2) the runtime memory footprint should fit comfortably within the available memory of the target device. Under this setting, all evaluated models satisfy the latency requirement, since even the slowest model required less than 20~ms per window. Memory is therefore the more restrictive factor. Models requiring more than approximately 1~GB of runtime memory, such as CLEF-Large, may be less suitable for low-power wearable or microcontroller-class deployment and would likely require compression, quantization, pruning, or offloading to a smartphone, gateway, or embedded GPU \citep{han2016deepcompression,sze2017efficient,googleLiteRTMicro}. Models with memory requirements below approximately 500~MB, such as ECGFounder, ECG-JEPA, ST-MEM, CLEF-Medium, CLEF-Small, and HuBERT-ECG-Small, are more realistic candidates for edge deployment on smartphones, embedded Linux platforms, or higher-end wearable processors, although further optimization may still be needed for ultra-low-power devices~\citep{googleLiteRTMicro}.

Overall, these results show that model scale alone does not determine deployment suitability. Larger models did not necessarily provide better AF detection performance, and in some cases introduced considerable memory and latency costs without improving accuracy. ECGFounder provided the best overall trade-off, combining the highest mean F\textsubscript{1}-score with moderate memory usage and acceptable inference time. This makes it the most promising candidate among the evaluated foundation models for practical AF screening in edge, bedside, and wearable monitoring scenarios.

\subsection{Performance Comparison with State-of-the-Art}
\label{subsec:sota}

Table~\ref{tab:sota_comparison} compares the performance of ECGFounder with previously reported state-of-the-art AF detection methods on the four benchmark datasets. It should be noted that the compared studies differ in terms of input representation, segment length, validation strategy, preprocessing pipeline, and classifier design. Therefore, the table should be interpreted as a contextual comparison rather than a strictly controlled head-to-head benchmark.
\begin{table*}[!t]
\caption{Cross-validation performance comparison between ECGFounder and state-of-the-art AF detection methods on the four benchmark datasets. Only cross-validation results are included. The highest value within each dataset is highlighted in bold. ``--'' indicates that the dataset was not evaluated or the metric was not reported. (BiLSTM: Bi-directional Long Short-Term Memory, AdaBoost: Adaptive Boosting, RRIs: R-R intervals, ViT: Vision Transformer)}
\label{tab:sota_comparison}
\centering
\setlength{\tabcolsep}{3.5pt}
\renewcommand{\arraystretch}{1.15}
\resizebox{\textwidth}{!}{%
\begin{tabular}{@{\extracolsep{8pt}}lcccccccccccccc}
\hline
\textbf{Reference} & \textbf{Method} & \textbf{Segment length}
& \multicolumn{3}{c}{\textbf{AFDB}} 
& \multicolumn{3}{c}{\textbf{CinC2017}} 
& \multicolumn{3}{c}{\textbf{CPSC2021}} 
& \multicolumn{3}{c}{\textbf{LTAFDB}} \\
\cline{4-6} \cline{7-9} \cline{10-12} \cline{13-15}
& & 
& Acc & Se & Sp 
& Acc & Se & Sp 
& Acc & Se & Sp 
& Acc & Se & Sp \\
\hline

Wen et al. \citep{wen2022comparative}  
& Stacked BiLSTM 
& 100 RRIs
& -- & -- & --
& -- & -- & --
& -- & -- & --
& 90.90 & 91.2  &  91.2 \\

Faust et al. \citep{faust2018automated}  
& BiLSTM
& 100 RRIs
& 98.51 & 98.32 & 98.67
& -- & -- & --
& -- & -- & --
& -- & -- & -- \\

Jahan et al. \citep{jahan2022short} 
& AdaBoost
& 20 RRIs
& 88.00 & 87.58 & 89.27
& -- & -- & --
& -- & -- & --
& -- & 86.45 & 81.57 \\

Duan et al. \citep{duan2022accurate} 
& RRI histogram + SVM
& 30 RRIs
& 98.43 & \textbf{98.48} & 98.40
& -- & -- & --
& -- & -- & --
& \textbf{98.40} & 95.24 & \textbf{99.94} \\

Pereira and Andre{\~a}o \citep{pereira2022inter} 
& RR features + LSTM
& 10~s
& 90.87 & 91.53 & 91.08
& -- & -- & --
& -- & -- & --
& 94.83  & 94.62 & 95.19 \\

Andersen et al. \citep{andersen2019deep} 
& CNN + LSTM
& 30 RRIs
&  97.10 & 98.17 & 96.29
& -- & -- & --
& -- & -- & --
& -- & -- & -- \\

Fan et al. \citep{fan2018multiscaled} 
& Multiscaled fusion of deep CNN
& 10~s
& -- & -- & --
& 97.72 & 94.31 & 98.22
& -- & -- & --
& -- & -- & -- \\

Zhang et al. \citep{zhang2023variable} 
& Time-adaptive densely network
& 10~s
& 87.40 & 88.00 & --
& \textbf{99.40} & \textbf{99.50} & --
& -- & -- & --
& -- & -- & -- \\

Sorayaie Azar et al. \citep{azar2025svit} 
& Spectrograms ViT 
& 7.8~s
& 95.97 & 89.43 & --
& 98.14 & 94.67 & --
& -- & -- & --
& -- & -- & -- \\

Hu et al. \citep{hu2023detection} 
& Residual blocks + Transformer encoder 
& 3 RRIs
& -- & -- & --
& -- & -- & --
& 98.15 & 98.06 & 98.31
& -- & -- & -- \\

Li et al. \citep{li2026robust} 
& Gated Contrastive Network 
& 10~s
& 98.01 & 97.94 & 98.06
& -- & -- & --
& 98.01 & 97.94 & 98.06
& -- & -- & -- \\

Zou et al. \citep{zou2024generalizable} 
& Residual CNN + BiLSTM 
& 30~s
& \textbf{98.63} & 97.35 & \textbf{99.49}
& -- & -- & --
& 98.63 & 99.08 & 98.03
& 97.04 & 96.81 & 97.31 \\

\hline

\textbf{ECGFounder}
& Frozen foundation model + XGBoost
& 10~s
& 98.33 & 97.04 & 99.11
& 98.15 & 92.86 & \textbf{98.92}
& \textbf{99.66} & \textbf{99.65} & \textbf{99.67}
& 95.49 & \textbf{96.81} & 94.27 \\

\hline

\end{tabular}%
}
\end{table*}

On AFDB, ECGFounder achieved an accuracy of 98.33\%, sensitivity of 97.04\%, and specificity of 99.11\%. Although Faust et al. \citep{faust2018automated} reported slightly higher accuracy and sensitivity using a BiLSTM model based on 100 RR intervals, ECGFounder achieved the highest specificity among the compared AFDB methods. Duan et al. \citep{duan2022accurate} also obtained strong performance using an RRI histogram and SVM, with accuracy, sensitivity, and specificity of 98.43\%, 98.48\%, and 98.40\%, respectively. These results indicate that ECGFounder performs competitively with both classical RRI-based approaches and deep learning models, while relying only on frozen foundation-model embeddings and a fixed downstream classifier. Importantly, ECGFounder uses shorter 10~s ECG windows, which typically contain approximately 10--20 RRIs depending on heart rate, compared to longer inputs of 100 RRIs and 30 RRIs in \citep{faust2018automated} and \citep{duan2022accurate}, which correspond roughly to 50--100~s and 15--30~s of ECG, respectively. Therefore, ECGFounder achieves competitive performance despite using substantially shorter temporal context.

For CinC2017, ECGFounder achieved 98.15\% accuracy, 92.86\% sensitivity, and 98.92\% specificity. Compared with \citep{fan2018multiscaled} and \citep{azar2025svit}, ECGFounder obtained the highest specificity, which indicates a lower false positive rate for AF detection in short single-lead ECG recordings. Zhang et al. \citep{zhang2023variable} reported higher accuracy and sensitivity. Overall, the results on CinC2017 suggest that ECGFounder provides robust discrimination in a challenging short-recording setting, particularly in terms of correctly identifying non-AF rhythms.

On CPSC2021, ECGFounder achieved the best overall performance among the compared methods, with 99.66\% accuracy, 99.65\% sensitivity, and 99.67\% specificity. The strong performance on CPSC2021 is consistent and confirms that ECGFounder provides highly discriminative representations for AF detection in this dataset. Notably, ECGFounder outperformed both hybrid and deep learning models specifically developed and trained for AF detection.

For LTAFDB, ECGFounder achieved the highest sensitivity of 96.81\%, which indicates strong capability in detecting AF windows in long-term ECG recordings. However, Duan et al. \citep{duan2022accurate} reported higher specificity and accuracy among the compared methods. This suggests that some RRI-based or hybrid models may be more conservative in reducing false positive detections on long-term Holter recordings. Nevertheless, ECGFounder provided the best sensitivity, which is clinically important for AF screening applications where missed AF episodes should be minimized.

Overall, ECGFounder achieved the best or highly competitive performance across the four datasets. In particular, it achieved the highest specificity on CinC2017, the best performance across all three metrics on CPSC2021, and the highest sensitivity on LTAFDB. These findings highlight the strong generalization capability of frozen ECG foundation model embeddings across heterogeneous datasets, while maintaining a simple and unified downstream classification pipeline.

\section{Discussion}
\label{sec:discussion}

The unified benchmark presented in this study provides a controlled comparison of the current generation of ECG foundation models for AF detection. Unlike the original publications of the evaluated models, which reported results using model-specific datasets, preprocessing procedures, downstream tasks, and validation settings, FOUND-AF evaluates all models under an identical and leakage-controlled protocol. By holding the preprocessing, downstream classifier, validation strategy, and decision threshold fixed, the framework reduces experimental confounding and enables performance differences to be attributed primarily to the quality of the pretrained ECG representations. Furthermore, the combined assessment of classification performance, statistical significance based on paired recording-level bootstrap comparisons with Holm correction, embedding separability, and computational efficiency provides evidence regarding not only predictive capability but also practical deployment feasibility.

The results also showed that model size alone does not determine downstream representation quality. For example, HuBERT-ECG-Base outperformed HuBERT-ECG-Large across all datasets despite having fewer parameters, and CLEF-Large did not outperform ECG-JEPA, ST-MEM, or ECGFounder despite being the largest model in the benchmark. These findings suggest that the relevance of the pretraining objective, the diversity and scale of the pretraining data, and the compatibility between pretraining and downstream tasks are more important than parameter count alone. This is particularly relevant for edge-oriented clinical applications, where larger models increase computational cost without necessarily improving diagnostic performance.

In addition, dataset-specific patterns were also observed. Performance was highest on CPSC2021, where the leading models achieved near-perfect F\textsubscript{1}-scores and AUC values. This may reflect the informative rhythm structure and signal characteristics of the dataset, although the large number of correlated windows within long recordings means that very high window-level performance should not be interpreted as direct evidence of event-level clinical performance. In contrast, CinC2017 was the most challenging dataset, likely due to its short single-lead recordings, low AF prevalence, and more variable signal quality. Several models achieved high specificity but substantially lower sensitivity on this dataset, which indicates a tendency to miss AF windows when only short rhythm fragments were available.

An important consideration in foundation model benchmarking is the possible influence of pretraining data overlap or pretraining-domain similarity. To reduce this risk, the four downstream datasets used in this study were selected carefully based on the available information about the pretraining corpora of the nine evaluated foundation models. To the best of our knowledge, AFDB, CinC2017, CPSC2021, and LTAFDB were not explicitly included in the pretraining datasets of these models. Nevertheless, several ECG foundation models are pretrained on large public or semi-public ECG corpora, and complete exclusion of overlap cannot be guaranteed unless detailed data origin and history are available. This does not necessarily imply data leakage, but it highlights the need for clearer documentation of pretraining datasets, exclusion criteria, and potential overlap with commonly used public benchmarks in future ECG foundation-model studies.

From a deployment perspective, the accuracy--efficiency analysis suggests that ECGFounder provides the most favorable trade-off among the evaluated models. It achieved the highest mean F\textsubscript{1}-score with 30.7 million parameters, 228~MB peak RSS memory, and 11.53~ms inference time per 10~s ECG window. Although ECG-JEPA and ST-MEM were faster, they required approximately twice the memory and achieved slightly lower classification performance. Conversely, larger models such as CLEF-Large and HuBERT-ECG-Large required substantially more memory without achieving competitive performance. These results indicate that compact, clinically relevant pretrained encoders may be more suitable for edge-supported AF screening than simply scaling model size. Nevertheless, deployment on ultra-low-power wearable or microcontroller-class devices would still require further optimization, such as quantization, pruning, compression, or offloading to a smartphone or embedded gateway.

Overall, the findings show that ECG foundation models can provide useful frozen representations for AF detection, but their performance depends strongly on pretraining strategy, dataset characteristics, and computational efficiency. ECGFounder emerged as the most promising model in this benchmark, combining robust classification performance with a practical resource profile for wearable, bedside, and edge-supported ECG monitoring.

\section{Conclusion}
\label{sec:conclusion}

This paper presented FOUND-AF, a unified and reproducible benchmark of nine publicly released ECG foundation models for AF detection across four heterogeneous ECG datasets. The evaluated models, chosen from the HuBERT-ECG, CLEF, ST-MEM, ECG-JEPA, and ECGFounder families, were assessed as frozen feature extractors on 1,954,594 ten-second ECG windows from 7,308 recordings. All experiments followed the same preprocessing, model-native resampling, grouped 5-fold cross-validation, and XGBoost downstream classification protocol, which enables a controlled comparison of the learned ECG representations.

The results identified ECGFounder as the strongest model across all datasets and evaluation metrics, achieving F\textsubscript{1}-scores of 97.87\%, 92.77\%, 99.50\%, and 95.68\% on AFDB, CinC2017, CPSC2021, and LTAFDB, respectively. Paired recording-level bootstrap analysis confirmed that its advantage was statistically significant in nearly all adjusted comparisons. The t-SNE, confusion-matrix, ROC, and efficiency analyses further showed that ECGFounder provides discriminative embeddings while maintaining a favourable accuracy--efficiency trade-off, with 30.7 million parameters, 228~MB peak RSS memory, and 11.53~ms inference time per 10~s window. These findings suggest that compact, clinically relevant ECG foundation models can support accurate AF detection without task-specific fine-tuning or on-device training.

The main limitations of this study are the restriction to binary AF-versus-Normal classification, the use of frozen encoders only, and the evaluation on publicly available datasets. Future work should extend the benchmark to multi-rhythm and multi-label arrhythmia detection, compare frozen and fine-tuned adaptation strategies, and validate the most promising models in prospective clinical, wearable, and edge-deployment settings. Overall, FOUND-AF provides a transparent reference for ECG foundation model evaluation and practical evidence that frozen pretrained encoders can support reliable AF detection under real-world computational constraints.

\bibliographystyle{unsrtnat}
\bibliography{references}  






\end{document}